\pdfoutput=1
\documentclass[11pt,a4paper]{article}
\usepackage[hyperref]{acl2021}
\usepackage{times}
\usepackage{latexsym}
\usepackage{subfig}

\usepackage{microtype}

\aclfinalcopy 
\def\aclpaperid{4049} 

\usepackage[T1]{fontenc}
\usepackage[utf8]{inputenc}
\usepackage[font=small,labelfont=bf]{caption}
\usepackage{enumitem}
\usepackage{soul}   
\usepackage{xcolor}

\usepackage{verbatim}

\usepackage{qtree}
\usepackage{graphics}

\usepackage{multirow}

\usepackage{microtype}

\title{
The Possible, the Plausible, and the Desirable:\\  
Event-Based Modality Detection for Language Processing}

\author{Valentina Pyatkin\thanks{~~Equal contribution}\\
 Bar Ilan University \\
  {\tt \footnotesize pyatkiv@biu.ac.il} \\
  \And
  {\bf Shoval Sadde\footnotemark[1]} \\
 Bar Ilan University \\
  {\tt\footnotesize shovatz@gmail.com} \\
  \And
  {\bf Aynat Rubinstein} \\
 Hebrew University of Jerusalem \\
  {\tt \footnotesize aynat.rubinstein@mail.huji.ac.il} \\
  \AND
 {\bf Paul Portner} \\
 Georgetown University\\
 {\tt \footnotesize paul.portner@georgetown.edu}\\
  \And
  {\bf Reut Tsarfaty} \\
Bar Ilan University \\

  {\tt \footnotesize reut.tsarfaty@biu.ac.il} 
 }

\begin{document}

\maketitle

\begin{abstract}
Modality is the linguistic ability to describe events with added information such as how {\em desirable, plausible}, or {\em feasible} they are.  Modality is important for many  NLP downstream tasks  such as the detection of hedging, uncertainty, speculation, and  more. 
Previous studies that address modality detection in NLP often restrict  modal expressions  to a closed \emph{syntactic class},  and the modal  {\em sense labels} are vastly different across different studies, lacking an accepted standard. Furthermore, these senses are often analyzed independently of the events  that they modify.  This work builds on the theoretical foundations  of the  {\em Georgetown Gradable Modal Expressions} (GME) work by \newcite{rubinstein2013toward} to propose an {\em event-based modality} detection  task where modal expressions can be words of any syntactic class and sense labels are drawn from a comprehensive taxonomy which harmonizes the modal concepts contributed by the different studies. 
We present experiments on the GME corpus aiming to detect  and classify fine-grained modal concepts and associate them with their  modified events. We show that detecting and classifying modal {expressions}   is not only feasible,  but also improves the detection of  {modal} {events} in their own right.
\end{abstract}

\section{Introduction}
 {\em Modality} refers to  the  linguistic ability to describe alternative ways the world could be.\footnote{In formal semantics, these alternatives are referred to as {\em possible worlds} or {\em situations} \citep{kripke1959,lewis1973,barwiseperry1981,kratzer2010}. } Modal expressions aim to identify wishes, rules, beliefs, or norms in texts \cite{Kratzer:1981,portner2009modality}, which is a crucial part of {\em Natural Language Understanding} (NLU)  \cite{morante-sporleder-2012-modality}. 
 
Concretely,  events in natural language are often reported in a manner that emphasizes {\em non-actual} perspectives on them, rather than their actual {\em propositional} content. Consider examples (1a)--(1b):
\begin{itemize}
        \item[(1)]
          \begin{itemize} 
          \item[a.] {\em We presented a paper at ACL'19.}
        \item[b.] {\em We did not present a paper at ACL'20.}
        \end{itemize}
        \end{itemize}
   The propositional content \(p=\)``present a paper at ACL'X''  can be easily verified for  sentences  (1a)-(1b) by looking up the proceedings of the conference to (dis)prove the existence of the relevant publication. The same proposition \(p\) is still referred to in sentences (2a)--(2d), but now in each one, \(p\)  is described from a different perspective:   
        \begin{itemize}
        \item[(2)]
    \begin{itemize}
    \item[a.] {\em We {\bf aim} to present a  paper at ACL'21.} 
    \item[b.] {\em We {\bf want} to  present a paper at ACL'21.}
   \item[c.] {\em We {\bf ought} to present a paper at ACL'21.}
\item[d.] {\em We are {\bf likely} to  present a paper at ACL'21.}
    \end{itemize}
\end{itemize}

These sentences cannot be verified or falsified simply by examining whether \(p\) {actually} came or will come to pass, and in fact, such verification is not  the goal of this way of reporting. Rather, speakers  describe  such events in order to indicate {\sc plans} (2a), {\sc desires} (2b), {\sc norms} (2c), or the assessed {\sc plausibility} (2d) of the associated propositional content \(p\). Investigating how to classify these perspectives on events has been the focus of extensive research on modality in theoretical linguistics \cite{Kratzer:1981,Palmer:1986,portner2009modality}.

In terms of NLP technology, modal concepts as  expressed in (2) are relevant to many downstream tasks,  such as the automatic detection of hedging and speculation \cite{vincze-08,malhotra-2013}, uncertainty \cite{vincze-08,miwa-12, zerva-17,prieto-2020}, opinion \cite{wiebe2005annotating, rubin-10, miwa-12}, and  factuality \cite{sauri2009, rudinger2018neural}. Although these tasks rely on modality features, so far there is no accepted standard for modal concepts and labels, which aligns with the semantic space of modal senses that linguists identify. Consequently, modality features are either treated idiosyncratically or are absent from  semantic frameworks \cite[\textsection 4.6]{donatelli-2018}.

In support of such downstream tasks, a different type of NLP investigations targets  modality annotation and detection  in its own right (\citet{ruppenhofer2012yes,baker2012modality,Zhouetal:2015,marasovic2016multilingual, hendrickx-12,nissim-2013,ghia-16,mendesetal2016modality, lavidetal2016linguistically}, and others). However, each of these studies creates its own scheme, and none of these schemes has been picked up as an accepted standard by the  community. Moreover,  different endeavors suffer from one (or more)  of the following types of deficiencies with respect to their expressivity and coverage. 

First, many studies limit the modal  {\em  triggers}, i.e., the expressions that trigger the modal meaning, to a closed class of auxiliary verbs (e.g., {\em can, might, should, must} in English \cite{ruppenhofer2012yes,marasovic2016modal,quersma-14}). However, as acknowledged by  linguists \cite{Kratzer:1981} and NLP researchers \cite{rubin-10,baker2012modality,nissim-2013}, words of any Part-of-Speech (POS)  can trigger modality. Consider, for instance, the following triggers: {\em We {\bf should} remain calm} (AUX); {\em We have a {\bf plan} to reduce the costs} (NOUN); {\em Our agency  {\bf prefers} this equipment} (VERB); {\em Marx is {\bf probably} patriotic} (ADV); {\em Devaluation has been {\bf necessary}} (ADJ).

Second, the  modal  {\em  senses}, i.e., the  {labels} that indicate the modal perspectives, differ from one  study to another, with no accepted standard. Some studies focus only on a particular sense,  such as epistemic modality \cite{rubin-10,ghia-16}. Others use labels that mix modal senses with orthogonal notions (e.g.,  {\em force}, distinguishing permission from requirement as in \citet{baker2012modality}), thereby making  their deployment into existing  annotations and  tasks less transparent.  In general, there is no single annotation standard that covers the full  spectrum of modal senses attested in the data and confirmed by the latest linguistic theories, as portrayed by \newcite{portner2009modality}. 

Finally, {\em modality detection} in NLP has often been cast as a  {word-sense disambiguation} (WSD) task \cite{ruppenhofer2012yes} or as a sentence-classification task \cite{marasovic2016multilingual}.  Both perspectives are insufficient for any  practical use. The latter is too coarse-grained, as a sentence may contain multiple events, each of which potentially carries a different modal sense. The former is uninformative, because the modal trigger is not explicitly associated with the event being modified. \newcite{ghia-16} take a step in the right direction, offering to annotate modal sense  {\em constructions}.

The current work proposes to address all of the aforementioned deficiencies as follows. We define a prediction task that we term {\em event-based modality detection}, where,  given a sentence as input, we aim to return  all of its modal {\em triggers}, their associated modal {\em senses}, and, for each trigger, the respective {\em event} being modified.  Crucially, the modal triggers can be from any syntactic class. The modal {senses} are drawn from a single taxonomy that we motivate based on linguistic research and which harmonizes the different modal concepts contributed in  previous studies (\textsection\ref{sec:taxonomy}). Finally, we propose to view modal triggers as semantic modifiers of eventive heads in  event-based (a.k.a., Neo-Davidsonian; \citet{parsons-90}) semantics. This  is motivated by practical concerns -- when extracting events from texts to benefit downstream tasks,  one would want easy access to the features that indicate the  perspective on each event, above and beyond  its participants.

The accompanying annotation standard we assume for the task is based on the {\em Georgetown Gradable Modal Expressions} (GME) framework \cite{rubinstein2013toward}, with  two simplifications  that are designed to allow for more consistent annotations and  increased ease-of-use by non-experts. First, we change the modal sense labels to be intuitive and  self-explanatory. Second, instead of the event span (a.k.a., {\em prejacent}) in the GME, we mark  the {\em head} of the event being modified.

To assess  the feasibility of the proposed task, we use the  GME corpus  \cite{rubinstein2013toward} to train and test the automatic detection of modal {\em triggers}, their {\em senses}, and associated {\em events}. Our experiments show that while identifying a closed set of auxiliary verbs as modal triggers is straightforward, expanding the set of triggers   to any syntactic class indeed makes it a harder task. Notwithstanding this difficulty, we show that a model based on large pre-trained contextualized embeddings  \cite{liu2019roberta} obtains substantial improvements over our baseline on the full task. Moreover, we show that detecting modalized events in fact improves with the availability of information about the modal triggers. All in all, we contribute a new task, a new standard and a set of strong baselines for the event-based modality task we defined. 

\section{Linguistic Background}
\label{sec:background}

Modal expressions allow language users to discuss alternative realities. For example, the sentence {\em She can reach the ceiling} is modal because it describes the event of her reaching the ceiling as feasible, but potentially non-actual. Similarly, {\em She hopefully will reach the ceiling} is modal because it describes such an event as desirable, and likewise potentially non-actual.   A sentence like {\em She was reported to reach the ceiling} describes the event of her reaching the ceiling as potentially actual, according to one's state of  knowledge, yet implying that in reality it could have been otherwise.

Over the last 40 years linguists have achieved an increasingly refined understanding of how to classify modal senses.  The most traditional and fundamental distinction is between {\em epistemic} modals and {\em non-epistemic} modals (also called {\em root} modals). Epistemic modals have to do with knowledge and plausibility of the event actually happening. Non-epistemic modals have to do with agent actions and motivations underlying the events.\footnote{The same split is motivated also on syntactic grounds: epistemic modals appear in high positions in the syntactic structure, in particular above tense and aspect, while root modals appear lower in the structure, closer to the verb phrase (see \citet{hacquard2010} for an overview).}

Epistemic modality is  not a unified class. Some modals express a perspective on the event that is based on knowledge, while others express a perspective related to the objective chance of the event happening  (a.k.a., {\em circumstantial} modality in \citet{Kratzer:1981}). Furthermore, linguists posit two types of {non-epistemic} modal senses: one which focuses on the {\em objective} abilities and  dynamic unfolding of events \citep{Palmer:1986}, and  another which focuses on {\em subjective} reasons to prioritise one event over another \citep{portner2009modality}. Within  the latter subtype there are further subdivisions according to whether the event is prioritised in terms of norms ({\em deontic}), desires/preferences ({\em bouletic}), or goals/plans ({\em teleological}) \citep{Kratzer:1981,portner2009modality, Rubinstein:PhD,MatthewsonTruckenbrodt:2018}.  

The traditional three-way classification of modal senses into {\em deontic, epistemic}, and {\em dynamic}, which has been used in previous NLP work (e.g., \citet{ruppenhofer2012yes,marasovic2016modal}), did not attend to these subdivisions, which are nonetheless expected to be important for reasoning and other tasks that require deep understanding. \newcite{baker2012modality} make finer-grained distinctions in the non-epistemic case, distinguishing between requirements, permissions, wants, and intentions, but not all of these in fact track distinct modal senses. For example, their ``require'' modality conflates both rule-based obligations and goal-oriented preferences. 

Most importantly, the discussion of modality in NLP often resorts to linguistic regimes that are not understandable by non-linguists and non-expert practitioners, making the output of these systems essentially unusable for NLP engineers and designers of downstream tasks. This paper aims to bridge this gap, offering a single task and annotation standard that cover the rich space of concepts, while being intuitively understandable and easy-to-use.

\paragraph{A Note on Modality vs.\ Factuality.}

A related but different line of work in NLP investigates the automatic identification and classification of the {\em factual} status of events \cite{sauri2009,rudinger2018neural}.  That is, the factuality classification task has to do with automatically detecting whether, in actuality, a reported event has {\em happened} or has {\em not happened}.\footnote{\newcite{rudinger2018neural} define factuality status on a scale of \{+3,-3\}. 0 indicates an event with unclear factuality status.}  

It is important to note that {\em factuality} and {\em modality} are distinct and completely orthogonal notions (see, e.g., \citealt{ghia-16}). For example, the sentences {\em The WSJ announced that she reached the shore} and  {\em She was able to reach the shore} share the propositional content of  \(p=\) `{\em she reached the shore}' and its implied factuality status (happened), but differ in the manner of reporting the event \(p\). The former is based on knowledge, while the latter puts emphasis on the ability of the agent in \(p\).  It is precisely this change of perspective that is missing in the realm of NLU and related downstream tasks. 

The upshot of \citeauthor{rudinger2018neural}'s (\citeyear{rudinger2018neural}) work is the claim that  factuality is determined at event level, and that expressions contributing to factuality may be of any syntactic class. We likewise propose to relate modal triggers to an event being modified, and we similarly adopt an inclusive view of the syntactic classes that express modality.  In contrast to event-based factuality detection, as proposed by \newcite{rudinger2018neural} and others, which classifies which events came to pass, event-based modality detection as  proposed here classifies an orthogonal dimension of meaning related to semantic properties of events that {\em may} be non-actual, providing information about {\em why} they are portrayed as such.

\section{Event-Based Modality Detection:  Proposed Task Definition}
\label{sec:taxonomy}
We propose an {\em event-based modality} detection  task that rests upon three  assumptions:  (i) the set of possible {\em modal triggers} is open-ended,  and may be of any POS tag, (ii) the associated {\em modal senses} are fine-grained and form an hierarchical taxonomy, and (iii) each  trigger is associated with an {\em event}. 

Consider, for instance, the following examples:
\begin{itemize}
\item[(3)]
\begin{itemize}
    \item[a.]  He was {\bf reported}\(_{i}\) to {\em be}\(_i\) in custody. 
    \item[b.] It is {\bf believed}\(_{j}\) that the glass will {\em make}\(_{j}\) it {\bf possible}\(_{k}\) to {\em see}\(_{k}\) the satellite at night.
    \end{itemize}
\end{itemize}
In these examples, the words in bold indicate the modal expression, which we call a \textit{trigger}. The co-indexed items in italics mark the {head} of the event for which the modal perspective is ascribed.  
In (3a), `{\bf reported}' triggers a modal perspective on the event of `{\em being (in custody)}'. In (3b), `{\bf believed}' triggers a modal perspective on the `{\em making}' event, and `{\bf possible}' indicates a modal perspective on the `{\em seeing (the satellite)}' event. 

Clearly, the modal perspectives on these events, i.e., the modal senses, are of different types. How should we label these fine-grained modal senses?

\paragraph{A Hierarchical Taxonomy of Modal Senses}

\begin{figure*}[t]
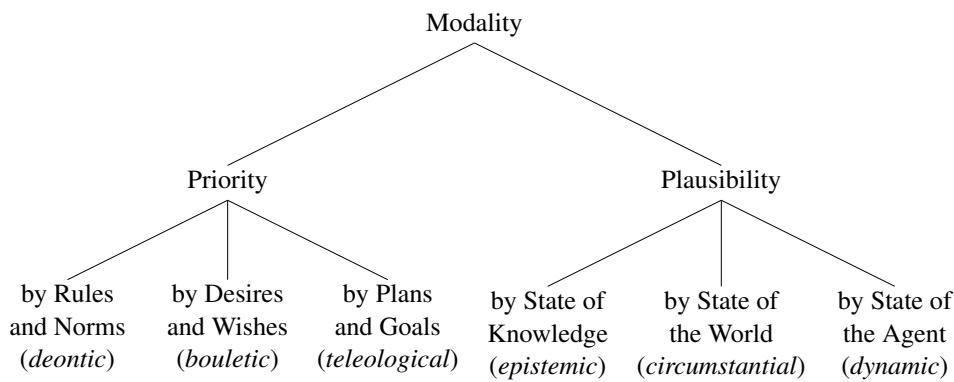

 \centering
 \scalebox{0.9}{
   \scalebox{1}{  \centering
 \Tree[.{Modality} [.Priority {by Rules}\\{and Norms}\\({\em deontic})     {by Desires}\\{and Wishes}\\({\em bouletic})   {by Plans}\\{and Goals}\\({\em teleological})  ]  [.Plausibility  {by State  of}\\{Knowledge}\\({\em epistemic})     {by State of}\\{the World}\\({\em circumstantial})  
 {by State of}\\{the Agent}\\({\em dynamic})
] ]    }}
     \caption{The Proposed Hierarchical Taxonomy of Modal Senses}
     \label{fig:tax}
 \end{figure*}

Having established that a given  expression  serves as a modal trigger, we are interested in classifying the  particular sense, or perspective, that it assigns to the modal event. Figure~\ref{fig:tax} presents the complete taxonomy that we propose for modal sense  classification in NLP. It is based on the modal senses proposed and justified by \newcite{rubinstein2013toward}, with a few simplifications that make it intuitive and easy-to-use by NLP practitioners and non-linguists.\footnote{Cf.\ Manning's Law, item 5 \url{https://en.wikipedia.org/wiki/Manning's_Law}}

The highest level of the hierarchy tracks the distinction between  events whose {\sc Plausibility} is being assessed, and events whose {\sc Priority} is stated. More specifically, {plausibility} has to do with events that are expected to happen or not happen, given a relevant set of assumptions which are made explicit. {Plausibility} can be assessed based on our state of knowledge (``I {\bf heard}\(_i\) she {\em got married}\(_i\)"), based on what is objectively  {probable} due to  facts about the world (``The ice cream will {\bf definitely}\(_i\) {\em melt}\(_i\) in the sun"), or based on  inherent (physical) abilities of an agent (``I {\bf can}\(_i\) easily {\em swim}\(_i\) 10 km").

In contrast, the {\sc Priority} branch marks a perspective where events  are {prioritized}, or considered ``good'' by the speaker (or more generally, by a relevant attitude holder) \cite{portner2009modality}. Events can be preferred because they are normatively obliged or commendable (``You {\bf should\(_i\)} n't {\em drink and drive}\(_i\)), because they realize a goal ("The {\bf plan}\(_i\) to {\em reduce}\(_i\)  costs in Q2"), or because they are otherwise desirable,  as a matter of personal taste or preference (``I will {\bf preferably}\(_i\) {\em meet}\(_i\) them over lunch'').

To make these notions accessible, we assign intuitive labels  to these fine-grained  concepts. On the {\sc Plausibility} side, we distinguish {plausibility} based on the state of {\sc knowledge} (previously, {\em epistemic}), plausibility based on a state of the {\sc world} ({\em circumstantial}), and plausibility based on the  objective abilities of the {\sc agent} ({\em dynamic}). On the {\sc Priority} side, we distinguish priorities based on {\sc rules and norms} ({\em deontic}), priorities based on {\sc desires and wishes} ({\em bouletic}), and  priorities based on  {\sc plans and goals} ({\em  teleological}). As illustrated in Table \ref{tab:pos-types}, modal triggers on both sides of the sense hierarchy may be of any POS tag.

\begin{table}[t]
    \centering
    \scalebox{0.8}{
\begin{tabular}{|l|l|}
\hline
\multicolumn{2}{|l|}{\textbf{Priority}} \\ \hline
Norms and Rules & \textit{\begin{tabular}[c]{@{}l@{}}the ballot which   {\bf must} be \\ held by the end of March\end{tabular}} \\ \hline
Desires and Wishes & \textit{\begin{tabular}[c]{@{}l@{}}we do   {\bf support} certain \\ limitations on the villains\end{tabular}} \\ \hline
Plans and Goals & \textit{\begin{tabular}[c]{@{}l@{}}a  {\bf necessity} emerged to \\ enter the Pilgrim's House\end{tabular}} \\ \hline
\multicolumn{2}{|l|}{\textbf{Plausibility}} \\ \hline
State of Knowledge & \textit{\begin{tabular}[c]{@{}l@{}}The ship is  {\bf believed} to \\ carry illegal immigrants\end{tabular}} \\ \hline
State of the World & \textit{\begin{tabular}[c]{@{}l@{}}The disease  {\bf can} be contr-\\ acted if a person is bitten\end{tabular}} \\ \hline
State of the Agent & \textit{\begin{tabular}[c]{@{}l@{}}They are  {\bf able}  to do \\ whatever they want\end{tabular}} \\ \hline
\end{tabular}}
    \caption{Modal-Sense Examples} 
    \label{tab:examples}
\end{table}

\begin{table*}[t]
    \centering
    \scalebox{0.85}{
    \begin{tabular}{|r|ll|}
    \hline
    & Priority & Plausibility \\
     \hline
       Aux  &   {\em We {\bf should} remain calm} & {\em there is little I {\bf can} do} \\
    \hline
       Verb    & {\em Our agency seriously {\bf needs} equipment} & {\em powers that {\bf enable} him to defend the rights}  \\
    
        Noun  & {\em a  {\bf plan} to reduce carbon-dioxide emissions}& {\em their {\bf incapacity} to put crime under control} \\
        Adverb   & {\em Marx is {\bf sufficiently} patriotic} & {\em President  Mugabe {\bf easily} won Zimbabwe's election} \\
        Adjective  & {\em devaluation was {\bf necessary}}&  {\em this complex decision was not {\bf easy} for him}\\
        \hline
    \end{tabular}}
    \caption{Modal Triggers with Diverse Parts-of-Speech Tags: Sentence Excerpts from the GME corpus.}
    \label{tab:pos-types}
\end{table*}

The proposed taxonomy unifies and harmonizes the different modal senses offered by previous studies.
Importantly, we enrich the {\em epistemic-deontic-dynamic} classification used in previous NLP research \cite{ruppenhofer2012yes,marasovic2016multilingual} with the finer-grained notions introduced by \newcite{rubinstein2013toward} and refer to the various labels in work by \newcite{baker2012modality,mendesetal2016modality}. More concretely, in GME and in our taxonomy, what in previous annotations was a monolithic {\em deontic} class \cite{ruppenhofer2012yes,marasovic2016multilingual} now corresponds to the {\sc Priority} node, with three linguistically-motivated sub-classes \cite{portner2009modality}: 
a {\sc rules-and-norms} class, a {\sc desires-and-wishes} class, and {\sc plans-and-goals}. 

Among modal events that do not involve priorities or norms, the sub-class which concerns the state of an {\sc agent} corresponds to {\em dynamic}   modality in previous studies \citep{ruppenhofer2012yes,marasovic2016modal}. The two other sub-classes of plausibility modality, state of {\sc world} and state of {\sc knowledge} taken together, correspond to {\em epistemic} in these previous works. 

To justify our fine-grained distinction, consider how the latter two senses, state of the {\sc world} and the state of {\sc knowledge}, correspond to interesting applications in the BioNLP literature, where it is vital to distinguish {\em fact} from {\em analysis} \citep{miwa-12}.  The difference is seen in the interpretations of {\bf may} in the following examples from the BioScope corpus \cite{vincze-08}:

\begin{itemize}
        \item[(4)]
          \begin{itemize} 
          \item[a.] {Symptoms {\bf may} {\em include} fever, cough or itches.} 
        \item[b.] {The presence of urothelial thickening and mild dilatation of the left ureter suggest that the patient {\bf may} {\em have} continued vesicoureteral reflux.} 
        \end{itemize}
\end{itemize}
In (4a), we classify {\bf may} to the {plausibility} branch with a state of the {\sc world} sub-class. 
In \citeauthor{miwa-12}'s terms this would be referred to as  {\em fact}. In (4b), we classify {\bf may} to the {plausibility} branch with a state of {\sc knowledge} sub-class. In \citeauthor{miwa-12}'s terms this would be referred to as  {\em analysis}.

\section{Experimental Setup}
\label{sec:exp}
\paragraph{Goal}
We set out to assess the feasibility of our proposed event-based modality task.
Concretely, we would like to gauge how well we can learn to detect and classify the different levels of modal senses afforded by our taxonomy (\textsection\ref{sec:taxonomy}) and to identify the  events modified by the  triggers.

\paragraph{Data}

Our experiments use the Georgetown Gradable Modal Expressions Corpus (GME; \citet{rubinstein2013toward}), a corpus obtained by expert annotations of the MPQA Opinion Corpus \cite{wiebe2005annotating}. The MPQA corpus is a 301,090-token corpus of news articles, which, following \newcite{ruppenhofer2012yes}, has become a benchmark for the annotation of modality. 

The GME corpus annotates various properties of modal expressions, including their sense in context, the proposition they apply to, the polarity of their environment, and whether or not they are qualified by a degree expression.\footnote{See \newcite{rubinstein2013toward} for details about the annotation process and the full scheme of annotated features.} \citet{rubinstein2013toward} claim inter-annotator agreement scores as follows: Krippendorf’s $\alpha = 0.89$ for a 2-way distinction corresponding to Priority versus Plausibility, $\alpha = 0.49$ for their finest-grained sense classification, and $\alpha = 0.65$ for prejacent span detection.

We processed the corpus by extracting the modal triggers and their corresponding propositional spans ({\em propositional argument} in GME) 
into a CoNLL-formatted file. Using spaCy \cite{spacy}, we obtained the lemmas, POS tags, and dependencies. The topmost head of the propositional span is considered the  head of the event being modified. We transformed the spans of modal propositions into  {\sc BIO}-tags, as shown in Table~\ref{tab:bio}.

We shuffled and split the data into 90\%  training and validation sets, and a 10\% test set. The training and validation set was then split into 5 folds, and in each fold, 20\% of the sentences were randomly assigned to validation, 80\% to training. As opposed to  \newcite{marasovic2016multilingual}, who trained and evaluated only on sentences already known to contain modal triggers,
we use the entire dataset, including sentences with no modality.\footnote{The processed data is available at \url{https://github.com/OnlpLab/Modality-Corpus}.}

\paragraph{Corpus Statistics} The GME corpus,  containing 11K sentences, shows that modality is a pervasive phenomenon (modal triggers were found in 96\% of the documents and in 48\% of the sentences). We  find  in the corpus 8318 modal triggers which correspond to 1502 unique types. 

Aside from verbs, nouns (e.g., {\em rights, possibility}) and adjectives (e.g., {\em fair, important}) are among the most frequently used modal expressions, with verbs making up 37\% of the modals in the corpus, adjectives 30\%,  and nouns 20\%. The remaining modals are either adverbials, auxiliaries, or particles. While most verbal triggers are modal verbs (e.g., {\em could, must, should}; {\sc mv} henceforth), 38\% have other POS tags.
736 triggers appear only once in the entire corpus with a modal meaning.\footnote{Words like  \textit{can} and {\em right} have non-modal meanings in addition to modal meanings.} 

About 25\% of modal triggers are ambiguous in terms of their modal sense (Plausibility vs. Priority), posing an additional classification challenge on top of the varied distribution of trigger POS tags. Modal triggers can also be multi-word expressions, with about 200 such instances in the corpus (e.g., {\em have to}). 

The modal-triggers' sense-labels are rather balanced: 48\% of the triggers in the corpus belong to ‘Plausibility’ and 52\% to ‘Priority’. For the finer-grained senses, the most common and least common classes make up 33\% and 7\% of the corpus, respectively.

\paragraph{The Proposed Tasks}
We experiment with three tasks, with an increasing level of complexity:  

 {\sc 1. Modal Sense Classification.} Here we aim to classify the modal sense of a trigger, assuming a modal trigger is already known. Specifically, we examine the contribution of the context to the lemma. We perform sense classification with the following variations:  (i) \textit{Vote}: a majority vote, (ii) \textit{Token}: out of context token-based classification where the trigger token is encoded using GloVe  \citep{pennington2014glove}), (iii) \textit{Context}: Token-in-context classification, given the whole sentence encoded with RoBERTa \cite{liu2019roberta} as input, with a marked trigger position, (iv) \textit{Masked}:  given the sentence  encoded with RoBERTa but with the trigger masked,
(v) \textit{Trigger+Head}:  only the trigger word and event head are given, encoded with RoBERTa, and finally, (vi) \textit{Full+Head}: the full sentence is encoded using RoBERTa with both the trigger and the event head marked. 

  {\sc 2. Modality Detection and Classification.}
This is a  realistic scenario, where we do not assume the trigger is known.
We aim to both identify the trigger and label its sense. We model this as a tagging task. Every token in the corpus is assigned a {\sc biose} tag  if it belongs to a modal trigger, which is appended with a suffix indicating its modal sense. We additionally perform variations of this task by including the head of the event as a feature (with either gold or predicted heads). Table~\ref{tab:bio} shows an example of the BIOSE tagging of modal triggers, with and without the event.

{\sc 3. Modal-Event Detection.}
Detecting and classifying modal triggers in isolation is insufficient for applications, as it is crucial to detect the event being modified. 
Here we predict a modal event and  aim to relate it to its trigger and modal sense.
We model this as sequence labeling, with the different tagging schemes to indicate the event being modified.
First, we aim to detect {\em only} the event.  In (i), we predict {\sc BIO} tags for the propositional spans.
   In (ii), we predict a {\sc Head} label  for  the event head.
   Next, we aim to {\em jointly} predict the modal  triggers and their modified events. To this end, 
    in (iii) we predict {\sc BIOSE-\{E|T\}} for the event span, concatenating the related modal trigger. That is, within a single event span marked with BIO, {\sc E} marks the propositional content and {\sc T} marks the  trigger. We experiment with and without  the {\em modal sense} appended to the trigger.
   Finally, in (iv) we predict  {\sc BIOSE}-\{sense\} tags that indicate the modal trigger along with a  {\sc Head}  tag for the event head.

\begin{table}[t]
 \centering
 \scalebox{0.77}{
\begin{tabular}{|l|l|l|l|}
\hline
\textbf{Text} & \textbf{BIOSE} & \textbf{Event Head} & \textbf{Event Span} \\ \hline \hline
Japan & O   &   O  &  O \\
has  & O   & O & O      \\
taken  & O   &  O  & O      \\
a  & O   & O & O     \\
leading        & O   &    O   & O     \\
role & O   &    O   & O     \\
in & O   &   O  & O     \\
the  & O   & O & O      \\
international  & O   &  O  & O     \\
drive  & S-{\sc Goals}  & S-{\sc Goals}  & B-T     \\
to & O   &   O  & I-E       \\
rebuild        & O   &    H    & I-E      \\
Afghanistan    & O   & O  & I-E      \\
... & O   & O & O      \\\hline

\end{tabular}}
\caption{Representing Event-Based Modality Using a {\sc bio} Tagging Scheme. On the left, the {\sc biose}-label tags are used to label the modal triggers. In the middle column {\sc bio} tags track the modal triggers, and H indicates the event head. On the right, the {\sc bio} tags track the event span, with the T and E labeling the trigger and event span respectively. }\label{tab:bio}
\end{table}

The labels that indicate modal sense are drawn from the proposed hierarchy, and we experiment with multiple levels of granularity: 
    {\em Modal/Not Modal}: a binary distinction,  indicating if the token is a modal trigger or not. {\em Coarse-grained}: a 3-way distinction, 
    indicating if the token is a modal trigger, and if so, what coarse-grained sense it has (Plausibility vs.\ Priority). {\em Fine-Grained}: indicating if the token is a modal trigger, and if so, which one of the senses at the lowest level of the hierarchy it has. We conflated {\em Desires/Wishes} and {\em Plans/Goals} into a single type called {\em Intentions}, since both these  senses are under-represented in our corpus. See appendix \ref{sec.data} for the  complete label distribution in our data.

\paragraph{Evaluation Metrics}
We report for all experiments {\sc biose}-chunk Precision, Recall, and (Macro) F1, calculated with the official \textit{ConllEval} script \cite{sang2000introduction}. 
When evaluating  span tagging for event-based modality we report labeled and unlabeled scores. 
When we report \textit{unlabeled} F1 for trigger classification, we check whether the token has been correctly identified as modal vs.\ not-modal, regardless of its sense.

\begin{table}[t]
    \centering
    \scalebox{0.77}{
    \begin{tabular}{|l||c c c c c c|}
    \hline
         & Vote & Token   & Mask & Context &  Head & Head  \\ 
          &  &    &  &   &  &  +Trigger  \\ \hline\hline
         Coarse & 89.1  & 88.7 & 78.0 & \textbf{90.7} & 90.5 & 90.1\\
         Fine & 72.0  & 72.4  & 58.3 & \textbf{76.4} & 76.2 & 75.1\\ \hline
    \end{tabular}}
    \caption{Modal Sense Classification with Oracle Triggers.} 
    \label{tab:senseclass}
\end{table}

\begin{table}[t]
    \centering
    \scalebox{0.77}{
    \begin{tabular}{|l|c c |c c|}
    \hline
    & Baseline &  & RoBERTa &  \\
         & {\sc mv}   & {\sc All}   & {\sc mv}   &  {\sc All}    \\
         \hline\hline
       Modal/Not  & 99.04 &  68.24 &  99.9 & 73.2 \\
        Coarse-Grained & 93.29 & 63.94 &  93.3 & 68.9 \\
        Fine-Grained  & 73.48 &  55.23 &  78.5 & 58.14 \\ \hline
    \end{tabular}}
    \caption{The Diversity of Modal Triggers: F1 of {\sc mv} triggers vs. All triggers, Majority Vote Baseline vs. RoBERTa}
    \label{tab:mvall}
\end{table}

\begin{table*}[t]
    \centering
    \scalebox{0.8}{
\begin{tabular}{|l l |ll l| l l l |l l l|}
\hline
\multicolumn{2}{|l|}{\multirow{2}{*}{}} & \multicolumn{3}{l|}{Modal/Not Modal} & \multicolumn{3}{l|}{Coarse-Grained} & \multicolumn{3}{l|}{Fine-Grained} \\ \cline{3-11} 
\multicolumn{2}{|l|}{} & P & R & F1 & P & R & F1 & P & R & F1 \\ \hline \hline
\multirow{2}{*}{Unlabeled} & Baseline & 75.81 & 62.07 & 68.24 & 75.81 & 62.07 & 68.24 & 75.81 & 62.07 & 68.24 \\ 
 & RoBERTa & 70.05 & 76.68 & 73.2 & 72.07 & 76.17 & 74.04 & 74.01 & 74.41 & 74.2 \\ \hline
\multirow{2}{*}{Labeled} & Baseline & NA & NA & NA & 71.36 & 57.92 & 63.94 & 58.68 & 45.56 & 51.29 \\  
 & RoBERTa & NA & NA & NA & 67.03 & 70.89 & 68.89 & 57.98 & 58.32 & 58.14 \\ \hline

\end{tabular}}
\caption{Precision, Recall, and F1 for Baseline and RoBERTa. In {\em labeled} the model tagged each token for modal/not modal and classified the identified modal tokens. In {\em unlabeled} the labels are given, but not counted beyond the modal/not-modal distinction.}
\label{tab:baseline-vs-roberta}
\end{table*}

\paragraph{Models}

Our baseline for modal trigger detection is a simple majority vote baseline where each token in the test set is tagged with its most frequent label in the training set.
For detecting modal triggers as well as for event detection,  we  experiment by fine-tuning a \textsc{RoBERTa}-based classifier \cite{liu2019roberta}.\footnote{We also experimented with a PyTorch-based sequence tagging model (NCRF++ by \citet{yang2018ncrfpp}) with  GoogleNews-vectors-negative300 embeddings (https://code.google.com/archive/p/word2vec/), but this setting did not outperform our majority vote baseline (and certainly under-performed the model based on contextualized representations),
and we didn't pursue this direction further.}  The encoded sequence is  fed through a linear layer with a softmax function predicting the appropriate tag for a given token. For the shorter spans (modal triggers) we predict the tag for every token-in-context. For the longer spans (events spans or events+trigger spans) we perform CRF decoding. The models we used are AllenNLP \cite{gardner2018allennlp} implementations. Whenever we use the trigger or the event as features to the model, we add special tokens to  the input, marking their respective spans in the sentence.
 {The hyperparameters of the models are as follows: we use $\textsc{RoBERTa}_{BASE}$ and fine-tune it for 6 epochs with a batch-size of 8, a learning rate of $1e^{-5}$ and the adam optimizer.}\footnote{The code for data processing, configuration files and training are available at \url{https://github.com/OnlpLab/Modality}.}

\section{Results}

\paragraph{Setting the Stage} 
Before evaluating our models  on the proposed tasks,  we first assess the empirical challenge of our {\em event-based modality detection} task relative to the modal sense {\em sentence classification} (SC) setup of \newcite{marasovic2016multilingual}.
Their work focuses on 6 modal auxiliary verbs ({\em can, could, may, must, should}, and {\em shall}) and modal senses from a restricted set of three labels ({\em deontic, dynamic, epistemic}).
Note that their proposed setup is not designed to separate modal sentences from non-modal ones, as the \citet{marasovic2016multilingual} dataset  contains {\em only} modal sentences. Second, it cannot directly indicate that a sentence contains multiple modal triggers with different senses. 

We trained and tested a CNN  compatible to theirs\footnote{Some dependencies  in the \newcite{marasovic2016multilingual} code are deprecated, so we use a simple off-the-shelf CNN model of AllenNLP \citep{gardner2018allennlp}.} on their  data  as well as our data (GME), using their proposed settings. We mapped our {\em Priority}, {\em Agent}, and {\em Knowledge} to their {\em deontic}, {\em dynamic}, and {\em epistemic}, respectively, and ignored our {\em State of the World (circumstantial)}. Here, we report the same sentence-based accuracy metrics as they do. 
Table~\ref{tab:replicate} shows the results on the two datasets, theirs 
and GME.
We see that accuracy on the SC task drops when switching from their data to ours, and that it drops  further when moving from a closed set of POS (Modal Verbs) to all targets.
All in all,  sentence classification is not sufficient to reflect the richness of {\em event-based} modality annotation, and  we conjecture that the SC setup would be too restrictive for real-world applications.

\begin{table}[t]
    \centering
    \scalebox{0.9}{
    \begin{tabular}{|r|c|}
    \hline
    Dataset - Triggers & Sentence Sense Accuracy \\
    \hline
    \hline
        Marasovi\`{c} - {\sc mv} &  79\\
        GME - {\sc mv} &  73\\
        GME - {\sc All} & 69 \\
        \hline
    \end{tabular}}
    \caption{Replicating the Setup of \newcite{marasovic2016multilingual} on the GME Data. Results drop for GME when using only sentences with modal verbs ({\sc MV}), and even further when using all of GME's sentences (namely with all modal triggers).}
    \label{tab:replicate}
\end{table}

\paragraph{Modal Sense Classification}
Next we report results for the first task we define, labeling the modal sense of an oracle trigger, as shown in Table \ref{tab:senseclass}. The majority vote baseline is high, which is partly due to the trigger lemma overlap between train and dev/test  (between 73\%-79\% depending on the split). Additionally we found only 25\% of the trigger lemmas  in the corpus to be ambiguous between Plausibility and Priority. Exposing the context, either by means of the full sentence or only the event head, improves results, and the improvement is more substantial for the fine-grained distinctions. Removing the lemma and using {\em only} context (Masked)  harms the results, but it is still impressive and shows that the environment has non-negligible contribution to sense disambiguation.
Finally, the sense classification is surprisingly effective also in cases where different modal events in the same sentence are intertwined.
An interesting example is the following sentence, with modal triggers in {\bf bold} (sense in brackets): "How \textbf{can}(Plausibility), under such circumstances, America \textbf{allow}(Priority) itself to express an \textbf{opinion}(Plausibility) over the issue of human \textbf{rights}(Priority) in other countries." Even when masking the triggers, the fine-tuned language model was able to correctly identify this alternating pattern of Plausibility and Priority.

\paragraph{Modal Triggers Detection}
Table \ref{tab:mvall} shows the modal trigger detection results  when applied only to the six modal verbs ({\sc mv}s), as opposed to modal triggers of unrestricted POS tags ({\sc All}). We see that when targeting only {\sc mv}s,  detecting modal elements is almost trivial for both the baseline and RoBERTa. Both models are also quite proficient (F1=93) at separating the different high-level modal senses (Priority vs.\ Plausibility) of the modal types that we defined.  
Once we switch to `All triggers', results substantially drop. 
Also, when switching to  finer-grained categories  we observe an expected drop for both the baseline and RoBERTa, with RoBERTa performing significantly better. 

Table \ref{tab:baseline-vs-roberta} presents the breakdown of the scores, labeled and unlabeled, for the different levels of granularity by the different models.  In all cases RoBERTa shows at least 5 absolute points consistent increase in F1 scores over the baseline, for all levels of granularity. Furthermore,   our unlabeled scores  demonstrate that predicting the  fine-grained categories by RoBERTa actually helps to determine the modal/non-modal  decision boundary, with an F1 improvement of about 1 absolute point at all levels. For the labeled accuracy, we observe an expected drop in the F1 scores when taking into account fine-grained labels. Yet, the performance is  better than a majority vote baseline and  is far better than chance for these nuanced distinctions. 

In the Fine-Grained Labeled RoBERTa setting the breakdown of the F1 performance by label is: \textit{agent}: 72.7, \textit{world}: 54.7, \textit{rules/norms}: 60.4, \textit{knowledge}: 59.3, \textit{intentional}: 46.1. These scores do not correlate with the frequency of each sense in the training data, e.g. \textit{agent} is the least frequent sense, but the model performed best at tagging it. Looking at ambiguous lemmas, i.e., lemmas that can have different modal senses depending on context, one can see that \textit{agent} and \textit{rules/norms} are the least ambiguous senses, which explains their higher performance scores. Breaking down the performance by coarse grained POS tag shows that \textsc{Verbs} are easiest to tag (66.5), followed by \textsc{Adverbs} (59.7), then \textsc{Adjectives} (55.9) and lastly, \textsc{Nouns}, which, with a score of 43.8, seem to be the hardest to tag. Interestingly, \textsc{Adjectives} are more ambiguous than \textsc{Nouns}; we thus do not have a satisfying explanation for why it is harder to classify the modality of \textsc{Noun} triggers.

\begin{table}[t]
    \centering
    \scalebox{0.77}{
    \begin{tabular}{|l||c c c c|}
    \hline
        F1 & No-Head & Head  & Head	 & Joint  
        \\ & & Gold & Predict & \\\hline\hline 
        Modal / Not-Modal      &  73.2 & 87.6 & 69.4  & 73.3 \\
         Coarse-Grained & 68.9 & 79.8  & 63.2 &  67.3 \\
         Fine-Grained &  58.14 & 66.7 & 52.1 &  56.0 \\ \hline
    \end{tabular}}
    \caption{Modal Trigger Tagging Results, F1 on Detected Spans, with and without Event Head Information.} 
    \label{tab:modaltagging}
\end{table}

Table \ref{tab:modaltagging} shows the  effect of event heads on modal trigger identification and classification, considering whether to model them separately or jointly in realistic scenarios, where the trigger is not known in advance. Gold event information as a feature for modal trigger tagging is helpful, but when this information is predicted,  propagated errors  decrease performance. Jointly predicting both triggers and event heads only very slightly decreases performance for the more fine-grained sense categories, making it a viable option for classification.

\paragraph{Event Detection Based on Modal Triggers}

Table~\ref{tab:prejacent} shows that event-span detection is  a harder task than merely locating the triggers (cf.\ Table~\ref{tab:baseline-vs-roberta}). Interestingly, predicting the span  {\em given} information about the trigger (Trigger as Feature) works  better than predicting the span with no such information (No-trigger). This holds both when the triggering event is provided by an Oracle (`Gold'), or whether it is predicted by RoBERTa (`Predict'). Improving modal trigger prediction is thus expected to further contribute to the accurate identification of events, and to event-span boundary detection. In general, head prediction shows better results than span prediction, 
partly due to the F1 score on spans being a  restrictive  metric in cases of partial overlap.

\paragraph{Error Analysis} To qualitatively assess the usability of RoBERTa's output, two trained human experts manually inspected the errors in 112 modal triggers in the dev set. Out of 36 false negatives (FN), 6 (16\% of the FN) are in fact correct (incorrectly tagged by the annotators as modal), and out of 27 false positives, 21 (78\% of the FP) are in fact correct (modals missed by the annotators). 
This leads to the conclusion that the gold annotation by the experts, while being precise, has incomplete coverage and lower recall. It implies  that RoBERTa's precision is in actuality {\em higher}, with a larger share of its predictions being correct. 

\begin{table}[t]
    \centering
    \scalebox{0.75}{
    \begin{tabular}{|l||l||c c c c|}
    \hline
        &F1 & No & Trigger   & Trigger  	 & Joint  \\ 
         &  & Trigger  & Gold   & Predict  	 & Joint  \\ \hline\hline
        \multirow{3}{*}{Span}& Modal / Not      &  51.1 & 71.13 & 53.55  & 50.05 \\
         & Coarse-Grained & 51.1 & 70.91 & 53.56 &  49.85 \\
         &Fine-Grained &  51.1 & 70.38 & 53.09 &  48.24 \\ \hline
          \multirow{3}{*}{Head}&Modal / Not      &  56.3 & 72.3 & 55.8  & 56.9 \\
         &Coarse-Grained & 56.3 & 71.6 & 56.0  &  60.7 \\
         &Fine-Grained &  56.3 & 70.9 & 55.2 & 55.3  \\ \hline
         
    \end{tabular}}
    \caption{Event Detection Results, F1 on Detected Spans, with and without Modal Trigger Information.}
    \label{tab:prejacent}
\end{table}

\section{Conclusion}
\label{sec:conclusion}

We propose an {\em event-based modality detection}  task which is based on solid theoretical foundations yet is adapted to fit  the needs of NLP practitioners. 
The task has three  facets: modal triggers can be of any syntactic type, sense labels are drawn from a unified taxonomy we propose, and modal triggers are  associated with their  modified  events. 
 We propose this task and standard as a potential extension for  standard semantic representations (AMR, SDG, UCCA, etc.) towards easy incorporation of modal events as features in downstream tasks. 

\section*{Acknowledgements}
We thank Yoav Goldberg, Ido Dagan, Noah Smith, Graham Katz, Elena Herburger, and members of the BIU-NLP Seminar for thoughtful feedback and fruitful discussion. We also thank 3 anonymous reviewers for their insightful remarks.
This research is supported by an ERC-StG grant of the European Research Council (no.\ 677352), the Israel Science Foundation (grant no. 1739/26 and grant no.\ 2299/19), and the National Science Foundation (BCS-1053038), for which we are grateful.

\bibliography{acl2021}

\begin{thebibliography}{40}
\expandafter\ifx\csname natexlab\endcsname\relax\def\natexlab#1{#1}\fi

\bibitem[{Baker et~al.(2012)Baker, Bloodgood, Dorr, Callison-Burch, Filardo,
  Piatko, Levin, and Miller}]{baker2012modality}
Kathryn Baker, Michael Bloodgood, Bonnie~J. Dorr, Chris Callison-Burch,
  Nathaniel~W. Filardo, Christine Piatko, Lori Levin, and Scott Miller. 2012.
\newblock \href {https://arxiv.org/pdf/1502.01682} {Use of modality and
  negation in semantically-informed syntactic {MT}}.
\newblock \emph{Computational Linguistics}, 38(2):411--438.

\bibitem[{Barwise and Perry(1981)}]{barwiseperry1981}
Jon Barwise and John Perry. 1981.
\newblock Situations and attitudes.
\newblock \emph{The Journal of Philosophy}, 78(11):668--691.

\bibitem[{Donatelli et~al.(2018)Donatelli, Regan, Croft, and
  Schneider}]{donatelli-2018}
Lucia Donatelli, Michael Regan, William Croft, and Nathan Schneider. 2018.
\newblock \href {https://www.aclweb.org/anthology/W18-4912} {Annotation of
  tense and aspect semantics for sentential {AMR}}.
\newblock In \emph{Proceedings of the Joint Workshop on Linguistic Annotation,
  Multiword Expressions and Constructions ({LAW}-{MWE}-{C}x{G}-2018)}, pages
  96--108, Santa Fe, New Mexico, USA. Association for Computational
  Linguistics.

\bibitem[{Gardner et~al.(2018)Gardner, Grus, Neumann, Tafjord, Dasigi, Liu,
  Peters, Schmitz, and Zettlemoyer}]{gardner2018allennlp}
Matt Gardner, Joel Grus, Mark Neumann, Oyvind Tafjord, Pradeep Dasigi, Nelson~F
  Liu, Matthew Peters, Michael Schmitz, and Luke Zettlemoyer. 2018.
\newblock \href {https://www.aclweb.org/anthology/W18-2501} {Allennlp: A deep
  semantic natural language processing platform}.
\newblock In \emph{Proceedings of Workshop for NLP Open Source Software
  (NLP-OSS)}, pages 1--6.

\bibitem[{Ghia et~al.(2016)Ghia, Kloppenburg, Nissim, and
  Pietrandrea}]{ghia-16}
Elisa Ghia, Lennart Kloppenburg, Malvina Nissim, and Paola Pietrandrea. 2016.
\newblock A construction-centered approach to the annotation of modality.
\newblock In \emph{Twelfth Joint ACL - ISO Workshop on Interoperable Semantic
  Annotation (ISA-12)}, pages 67--74, Portorož, Slovenia.

\bibitem[{Hacquard(2010)}]{hacquard2010}
Valentine Hacquard. 2010.
\newblock On the event relativity of modal auxiliaries.
\newblock \emph{Natural Language Semantics}, 18:79--114.

\bibitem[{Hendrickx et~al.(2012)Hendrickx, Mendes, and
  Mencarelli}]{hendrickx-12}
Iris Hendrickx, Amália Mendes, and Silvia Mencarelli. 2012.
\newblock Modality in text: a proposal for corpus annotation.
\newblock In \emph{Proceedings of the Eight International Conference on
  Language Resources and Evaluation (LREC'12)}, Istanbul, Turkey. European
  Language Resources Association (ELRA).

\bibitem[{Honnibal et~al.(2020)Honnibal, Montani, Van~Landeghem, and
  Boyd}]{spacy}
Matthew Honnibal, Ines Montani, Sofie Van~Landeghem, and Adriane Boyd. 2020.
\newblock \href {https://doi.org/10.5281/zenodo.1212303} {{spaCy:
  Industrial-strength Natural Language Processing in Python}}.

\bibitem[{Kratzer(1981)}]{Kratzer:1981}
Angelika Kratzer. 1981.
\newblock \href {http://users.ox.ac.uk/~sfop0776/KratzerNotional.pdf} {The
  notional category of modality}.
\newblock In Hans-J{\"u}rgen Eikmeyer and Hannes Rieser, editors, \emph{Words,
  Worlds, and Contexts}, pages 38--74. Walter de Gruyter, Berlin.
\newblock Reprinted in \emph{Formal Semantics: The Essential Readings}, ed.
  Paul Portner and Barbara H. Partee (2002), 289--323. Oxford: Blackwell.

\bibitem[{Kratzer(2010)}]{kratzer2010}
Angelika Kratzer. 2010.
\newblock \href
  {http://plato.stanford.edu/archives/fall2010/entries/situations-semantics/}
  {Situations in natural language semantics}.
\newblock In Edward~N. Zalta, editor, \emph{The {S}tanford Encyclopedia of
  Philosophy}, fall 2010 edition.
\newblock First published February 2007.

\bibitem[{Kripke(1959)}]{kripke1959}
Saul~A. Kripke. 1959.
\newblock A completeness theorem in modal logic.
\newblock \emph{The Journal of Symbolic Logic}, 24(1):1--14.

\bibitem[{Lavid et~al.(2016)Lavid, Carretrero, and
  Zamorano-Mansilla}]{lavidetal2016linguistically}
Julia Lavid, Marta Carretrero, and Juan~Rafael Zamorano-Mansilla. 2016.
\newblock \href {https://www.aclweb.org/anthology/2016.lilt-14.4} {A
  linguistically-motivated annotation model of modality in {E}nglish and
  {S}panish: Insights from {MULTINOT}}.
\newblock In \emph{Linguistic Issues in Language Technology, Volume 14, 2016 -
  Modality: Logic, Semantics, Annotation, and Machine Learning}. CSLI
  Publications.

\bibitem[{Lewis(1973)}]{lewis1973}
David Lewis. 1973.
\newblock \emph{Counterfactuals}.
\newblock Harvard University Press, Cambridge, Mass.

\bibitem[{Liu et~al.(2019)Liu, Ott, Goyal, Du, Joshi, Chen, Levy, Lewis,
  Zettlemoyer, and Stoyanov}]{liu2019roberta}
Yinhan Liu, Myle Ott, Naman Goyal, Jingfei Du, Mandar Joshi, Danqi Chen, Omer
  Levy, Mike Lewis, Luke Zettlemoyer, and Veselin Stoyanov. 2019.
\newblock \href {https://arxiv.org/pdf/1907.11692} {Ro{BERT}a: A robustly
  optimized {BERT} pretraining approach}.
\newblock \emph{arXiv preprint arXiv:1907.11692}.

\bibitem[{Malhotra et~al.(2013)Malhotra, Younesi, Gurulingappa, and
  Hofmann-Apitius}]{malhotra-2013}
Ashutosh Malhotra, Erfan Younesi, Harsha Gurulingappa, and Martin
  Hofmann-Apitius. 2013.
\newblock \href {https://doi.org/10.1371/journal.pcbi.1003117}
  {`{HypothesisFinder}:' a strategy for the detection of speculative statements
  in scientific text}.
\newblock \emph{PLOS Computational Biology}, 9(7):e1003117.

\bibitem[{Marasovi{\'c} and Frank(2016)}]{marasovic2016multilingual}
Ana Marasovi{\'c} and Anette Frank. 2016.
\newblock \href {https://www.aclweb.org/anthology/W/W16/W16-1613.pdf}
  {Multilingual modal sense classification using a convolutional neural
  network}.
\newblock In \emph{Proceedings of the 1st Workshop on Representation Learning
  for NLP}, pages 111--120.

\bibitem[{Marasovi{\'c} et~al.(2016)Marasovi{\'c}, Zou, Palmer, and
  Frank}]{marasovic2016modal}
Ana Marasovi{\'c}, Mengfei Zou, Alexis Palmer, and Anette Frank. 2016.
\newblock \href {https://www.aclweb.org/anthology/2016.lilt-14.3/} {Modal sense
  classification at large. paraphrase-driven sense projection, semantically
  enriched classification models and cross-genre evaluations}.
\newblock \emph{LiLT (Linguistic Issues in Language Technology)}, 14.

\bibitem[{Matthewson and Truckenbrodt(2018)}]{MatthewsonTruckenbrodt:2018}
Lisa Matthewson and Hubert Truckenbrodt. 2018.
\newblock \href
  {https://www.leibniz-zas.de/fileadmin/Archiv2019/mitarbeiter/truckenbrodt/2018_LM_HT.pdf}
  {Modal flavour/modal force interactions in {G}erman: {\em soll}, {\em
  sollte}, {\em muss} and {\em m\"usste}}.
\newblock \emph{Linguistische Berichte}, 255:259--312.

\bibitem[{Mendes et~al.(2016)Mendes, Hendrickx, {\'A}vila, Quaresma,
  Gon\c{c}alves, and Sequeira}]{mendesetal2016modality}
Am{\'a}lia Mendes, Iris Hendrickx, Liciana {\'A}vila, Paulo Quaresma, Teresa
  Gon\c{c}alves, and Jo{\~a}o Sequeira. 2016.
\newblock \href {https://www.aclweb.org/anthology/2016.lilt-14.5} {Modality
  annotation for {P}ortuguese: from manual annotation to automatic labeling}.
\newblock In \emph{Linguistic Issues in Language Technology, Volume 14, 2016 -
  Modality: Logic, Semantics, Annotation, and Machine Learning}. CSLI
  Publications.

\bibitem[{Miwa et~al.(2012)Miwa, Thompson, McNaught, Kell, and
  Ananiadou}]{miwa-12}
Makoto Miwa, Paul Thompson, John McNaught, Douglas~B. Kell, and Sophia
  Ananiadou. 2012.
\newblock \href {https://doi.org/10.1186/1471-2105-13-108} {Extracting
  semantically enriched events from biomedical literature}.
\newblock \emph{BMC Bioinformatics}, 13(108).

\bibitem[{Morante and Sporleder(2012)}]{morante-sporleder-2012-modality}
Roser Morante and Caroline Sporleder. 2012.
\newblock \href {https://doi.org/10.1162/COLI_a_00095} {Modality and negation:
  An introduction to the special issue}.
\newblock \emph{Computational Linguistics}, 38(2):223--260.

\bibitem[{Nissim et~al.(2013)Nissim, Pietrandrea, Sans{\`o}, and
  Mauri}]{nissim-2013}
Malvina Nissim, Paola Pietrandrea, Andrea Sans{\`o}, and Caterina Mauri. 2013.
\newblock \href {https://www.aclweb.org/anthology/W13-0501} {Cross-linguistic
  annotation of modality: a data-driven hierarchical model}.
\newblock In \emph{Proceedings of the 9th Joint {ISO} - {ACL} {SIGSEM} Workshop
  on Interoperable Semantic Annotation}, pages 7--14, Potsdam, Germany.
  Association for Computational Linguistics.

\bibitem[{Palmer(1986)}]{Palmer:1986}
Frank~R. Palmer. 1986.
\newblock \href {https://apps.dtic.mil/dtic/tr/fulltext/u2/a400315.pdf}
  {\emph{Mood and Modality}}.
\newblock Cambridge University Press, Cambridge.

\bibitem[{Parsons(1990)}]{parsons-90}
Terence Parsons. 1990.
\newblock \emph{Events in the Semantics of {English}: A Study in Subatomic
  Semantics}.
\newblock MIT Press, Cambridge, MA.

\bibitem[{Pennington et~al.(2014)Pennington, Socher, and
  Manning}]{pennington2014glove}
Jeffrey Pennington, Richard Socher, and Christopher~D Manning. 2014.
\newblock Glove: Global vectors for word representation.
\newblock In \emph{Proceedings of the 2014 conference on empirical methods in
  natural language processing (EMNLP)}, pages 1532--1543.

\bibitem[{Portner(2009)}]{portner2009modality}
Paul Portner. 2009.
\newblock \emph{Modality}.
\newblock Oxford University Press.

\bibitem[{Prieto et~al.(2020)Prieto, Deus, de~Waard, Schultes,
  García-Jiménez, and Wilkinson}]{prieto-2020}
Mario Prieto, Helena Deus, Anita de~Waard, Erik Schultes, Beatriz
  García-Jiménez, and Mark~D. Wilkinson. 2020.
\newblock \href {https://doi.org/10.7717/peerj.8871} {Data-driven
  classification of the certainty of scholarly assertions}.
\newblock \emph{PeerJ 8}, 8:e8871.

\bibitem[{Quaresma et~al.(2014)Quaresma, Mendes, Hendrickx, and
  Gon\c{c}alves}]{quersma-14}
Paulo Quaresma, Amália Mendes, Iris Hendrickx, and Teresa Gon\c{c}alves. 2014.
\newblock Automatic tagging of modality: identifying triggers and modal value.
\newblock In \emph{The 10th Joint ACL SIGSEM - ISO Workshop on Interoperable
  Semantic Annotation}, pages 95--102.

\bibitem[{Rubin(2010)}]{rubin-10}
Victoria~L. Rubin. 2010.
\newblock \href
  {http://www.sciencedirect.com/science/article/pii/S0306457310000208}
  {Epistemic modality: From uncertainty to certainty in the context of
  information seeking as interactions with texts}.
\newblock \emph{Information Processing \& Management}, 46(5):533--540.

\bibitem[{Rubinstein(2012)}]{Rubinstein:PhD}
Aynat Rubinstein. 2012.
\newblock \emph{Roots of Modality}.
\newblock Ph.D. thesis, University of Massachusetts Amherst.

\bibitem[{Rubinstein et~al.(2013)Rubinstein, Harner, Krawczyk, Simonson, Katz,
  and Portner}]{rubinstein2013toward}
Aynat Rubinstein, Hillary Harner, Elizabeth Krawczyk, Dan Simonson, Graham
  Katz, and Paul Portner. 2013.
\newblock \href {https://www.aclweb.org/anthology/W13-0306} {Toward
  fine-grained annotation of modality in text}.
\newblock In \emph{Proceedings of the IWCS 2013 Workshop on Annotation of Modal
  Meanings in Natural Language (WAMM)}, pages 38--46.

\bibitem[{Rudinger et~al.(2018)Rudinger, White, and
  Van~Durme}]{rudinger2018neural}
Rachel Rudinger, Aaron~Steven White, and Benjamin Van~Durme. 2018.
\newblock \href {https://doi.org/10.18653/v1/N18-1067} {Neural models of
  factuality}.
\newblock In \emph{Proceedings of the 2018 Conference of the North {A}merican
  Chapter of the Association for Computational Linguistics: Human Language
  Technologies, Volume 1 (Long Papers)}, pages 731--744, New Orleans,
  Louisiana. Association for Computational Linguistics.

\bibitem[{Ruppenhofer and Rehbein(2012)}]{ruppenhofer2012yes}
Josef Ruppenhofer and Ines Rehbein. 2012.
\newblock \href {https://www.aclweb.org/anthology/L12-1458/} {Yes we can!?
  annotating english modal verbs}.
\newblock In \emph{Proceedings of the Eighth International Conference on
  Language Resources and Evaluation (LREC-2012)}, pages 1538--1545.

\bibitem[{Sang and Buchholz(2000)}]{sang2000introduction}
Erik Tjong~Kim Sang and Sabine Buchholz. 2000.
\newblock Introduction to the conll-2000 shared task chunking.
\newblock In \emph{Fourth Conference on Computational Natural Language Learning
  and the Second Learning Language in Logic Workshop}.

\bibitem[{Saur\'i and Pustejovsky(2009)}]{sauri2009}
Roser Saur\'i and James Pustejovsky. 2009.
\newblock \href
  {https://www.academia.edu/download/45932196/FactBank_a_corpus_annotated_with_event_f20160524-28223-hblb2r.pdf}
  {{FactBank}: a corpus annotated with event factuality}.
\newblock \emph{Language Resources and Evaluation}, 43:227--268.

\bibitem[{Vincze et~al.(2008)Vincze, Szarvas, Farkas, Móra, and
  Csirik}]{vincze-08}
Veronika Vincze, György Szarvas, Richárd Farkas, György Móra, and János
  Csirik. 2008.
\newblock \href {https://doi.org/10.1186/1471-2105-9-S11-S9} {The {BioScope}
  corpus: biomedical texts annotated for uncertainty, negation and their
  scopes}.
\newblock \emph{BMC Bioinformatics}, 9:S9.

\bibitem[{Wiebe et~al.(2005)Wiebe, Wilson, and Cardie}]{wiebe2005annotating}
Janyce Wiebe, Theresa Wilson, and Claire Cardie. 2005.
\newblock \href {https://link.springer.com/article/10.1007/s10579-005-7880-9}
  {Annotating expressions of opinions and emotions in language}.
\newblock \emph{Language Resources and Evaluation}, 39(2-3):165--210.

\bibitem[{Yang and Zhang(2018)}]{yang2018ncrfpp}
Jie Yang and Yue Zhang. 2018.
\newblock \href {https://www.aclweb.org/anthology/P18-4013} {{NCRF}++: An
  open-source neural sequence labeling toolkit}.
\newblock \emph{arXiv preprint arXiv:1806.05626}.

\bibitem[{Zerva et~al.(2017)Zerva, Batista-Navarro, Day, and
  Ananiadou}]{zerva-17}
Chrysoula Zerva, Riza Batista-Navarro, Philip Day, and Sophia Ananiadou. 2017.
\newblock \href {doi: 10.1093/bioinformatics/btx466} {Using uncertainty to link
  and rank evidence from biomedical literature for model curation}.
\newblock \emph{Bioinformatics}, 33(23):3784--3792.

\bibitem[{Zhou et~al.(2015)Zhou, Frank, Friedrich, and Palmer}]{Zhouetal:2015}
Mengfei Zhou, Anette Frank, Annemarie Friedrich, and Alexis Palmer. 2015.
\newblock \href {https://doi.org/10.18653/v1/W15-2705} {Semantically enriched
  models for modal sense classification}.
\newblock In \emph{Proceedings of the First Workshop on Linking Computational
  Models of Lexical, Sentential and Discourse-level Semantics}, pages 44--53,
  Lisbon, Portugal. Association for Computational Linguistics.

\end{thebibliography}
\bibliographystyle{acl_natbib}

\clearpage
\appendix













\appendix\section{Data}\label{sec.data} 
\subsection{GME in numbers}
The GME dataset \cite{rubinstein2013toward} annotates the MPQA corpus \citep{wiebe2005annotating} with information about modality. The corpus consists of 534 documents which in turn contain 11,048 sentences. 5288 sentences have modal triggers, and of them, in 1141 the modal trigger is an  auxiliary verb. There are 7979 instances of modal triggers (tokens), which belong to 1141 unique words (types). 1229 of the modal triggers are modal verbs.
The breakdown of the modal triggers into the different modal senses is given in Table \ref{tab:counts}.

\begin{table}[h!]
    \centering
   \scalebox{0.85}{
\begin{tabular}{|c||c|c|c|c|}
\hline
Type & Quantity & \multicolumn{2}{c|}{\begin{tabular}[c]{@{}c@{}}2-way \\ ambiguity\end{tabular}} & \begin{tabular}[c]{@{}c@{}}3-way\\ ambiguity\end{tabular} \\ \hline
    \hline
\begin{tabular}[c]{@{}c@{}}Rules \& \\  Norms\end{tabular} & 2316 & \multicolumn{2}{c|}{} & \multirow{3}{*}{537} \\ \cline{1-4}
\begin{tabular}[c]{@{}c@{}}Desires \&\\  Wishes\end{tabular} & 142 & \multicolumn{2}{c|}{\multirow{2}{*}{210}} &  \\ \cline{1-2}
\begin{tabular}[c]{@{}c@{}}Plans \&\\  Goals\end{tabular} & 1077 & \multicolumn{2}{c|}{} &  \\ \hline
Knowledge & 1527 & \multirow{2}{*}{557} &  & \multirow{3}{*}{} \\ \cline{1-2} \cline{4-4}
World & 1303 &  & \multirow{2}{*}{202} &  \\ \cline{1-3}
Agent & 447 &  &  &  \\ \hline
\end{tabular}}
    \caption{Label Counts in the GME Data}
    \label{tab:counts}
\end{table}

\subsection{Data Pre-processing}
We parsed the data using spaCy, and obtained the lemma, POS, and dependency information for all tokens in our corpus.
We split the data into 5 folds, where each fold had a different split of training and validation set, but the test set is the same for all folds.
Train and validation sets are of 9894 sentences (validation 1975 and training 7919), while the test set has 1096 sentences.
The train and validation sets have 7160 modal triggers, while the test set has 819.

\section{Additional Materials}
Please refer to the following github repositories for code and data:
\paragraph{Code} Code and models and evaluation scripts used in our experiments

\url{https://github.com/OnlpLab/Modality}
\paragraph{Data} A processed version of the GME corpus, including all annotation layers and meta-information. 

\url{https://github.com/OnlpLab/Modality-Corpus}

\section{Experimental Setting}

We had 4 GeForce GTX 1080 Ti available for training and hyper-parameter search.  Our models are based on RoBERTa-base, which has 82M parameters and it takes about 45 minutes to train a single tagging model.

Tables \ref{tab:baseline} and \ref{tab:roberta} show the results of the baseline and RoBERTa respectively. On the right hand side of the tables, the scores are split by modal senses.
Here too, we observe that RoBERTa obtains substantial improvements on  per-label scores over the baseline.

\begin{table*}[h!]
    \centering
    \scalebox{1}{
\begin{tabular}{|c|c|c|c|c|c|c|c|c|}
\hline
 & \multicolumn{2}{c|}{\textbf{P/R/F1}} & \multicolumn{6}{c|}{\textbf{Labels F1}} \\ \hline
 & \textbf{Labeled} & \textbf{Unlabeled} & \multicolumn{3}{c|}{\textbf{}} & \multicolumn{3}{c|}{\textbf{}} \\ \hline\hline
\multirow{3}{*}{\textbf{\begin{tabular}[c]{@{}c@{}}Modal vs. \\ Not-Modal\end{tabular}}} & 75.81 & 75.81 & \multicolumn{3}{c|}{\multirow{3}{*}{}} & \multicolumn{3}{c|}{\multirow{3}{*}{}} \\
 & 62.07 & 62.07 & \multicolumn{3}{c|}{} & \multicolumn{3}{c|}{} \\
 & 68.24 & 68.24 & \multicolumn{3}{c|}{} & \multicolumn{3}{c|}{} \\ \hline
\multirow{3}{*}{\textbf{\begin{tabular}[c]{@{}c@{}}Priority vs. \\ Plausibility\end{tabular}}} & 71.36 & 75.81 & \multicolumn{3}{c|}{\textbf{Priority}} & \multicolumn{3}{c|}{\textbf{Plausibility}} \\
 & 57.92 & 62.07 & \multicolumn{3}{c|}{\multirow{2}{*}{55.46}} & \multicolumn{3}{c|}{\multirow{2}{*}{72.51}} \\
 & 63.94 & 68.24 & \multicolumn{3}{c|}{} & \multicolumn{3}{c|}{} \\ \hline
\multirow{3}{*}{\textbf{Fine-Grained}} & 58.68 & 75.81 & \textbf{Rules} & \multicolumn{2}{c|}{\textbf{Intentions*}} & \textbf{Knowledge} & \textbf{World} & \textbf{Agent} \\
 & 45.56 & 62.07 & \multirow{2}{*}{50.94} & \multicolumn{2}{c|}{\multirow{2}{*}{39.11}} & \multirow{2}{*}{50.95} & \multirow{2}{*}{52.58} & \multirow{2}{*}{67.39} \\
 & 51.29 & 68.24 &  & \multicolumn{2}{c|}{} &  &  &  \\ \hline

\end{tabular}
}
    \caption{Classifying Modal Events: Baseline Results (ambiguities not shown). We unified wishes and goals into \textit{intentions} for reasons of data sparsity.}
    \label{tab:baseline}

\vspace{2cm}
    \centering
    \scalebox{1}{
    \begin{tabular}{|c|c|c|||c|c|c|c|c|c|}
\hline
 & \multicolumn{2}{c|}{\textbf{P/R/F1}} & \multicolumn{6}{c|}{\textbf{Labeled F1}} \\ \hline
 & \textbf{Labeled} & \textbf{Unlabeled} & \multicolumn{3}{c|}{\textbf{}} & \multicolumn{3}{c|}{\textbf{}} \\ \hline
 \hline
\multirow{3}{*}{\textbf{\begin{tabular}[c]{@{}c@{}}Modal vs. \\ Not-Modal\end{tabular}}} & NA & 70.05 & \multicolumn{3}{c|}{\multirow{3}{*}{}} & \multicolumn{3}{c|}{\multirow{3}{*}{}} \\
 & NA & 76.68 & \multicolumn{3}{c|}{} & \multicolumn{3}{c|}{} \\
 & NA & 73.2 & \multicolumn{3}{c|}{} & \multicolumn{3}{c|}{} \\ \hline
\multirow{3}{*}{\textbf{\begin{tabular}[c]{@{}c@{}}Priority vs. \\ Plausibility\end{tabular}}} & 67.03 & 72.07 & \multicolumn{3}{c|}{\textbf{Priority}} & \multicolumn{3}{c|}{\textbf{Plausibility}} \\
 & 70.89 & 76.17 & \multicolumn{3}{c|}{\multirow{2}{*}{62.98}} & \multicolumn{3}{c|}{\multirow{2}{*}{75.52}} \\
 & 68.89 & 74.04 & \multicolumn{3}{c|}{} & \multicolumn{3}{c|}{} \\ \hline
\multirow{3}{*}{\textbf{Fine-Grained}} & 57.98 & 74.01 & \textbf{Rules} & \multicolumn{2}{c|}{\textbf{Intentions*}} & \textbf{Knowledge} & \textbf{World} & \textbf{Agent} \\
 & 58.32 & 74.41 & \multirow{2}{*}{60.42} & \multicolumn{2}{c|}{\multirow{2}{*}{46.1}} & \multirow{2}{*}{59.27} & \multirow{2}{*}{54.64} & \multirow{2}{*}{72.72} \\
 & 58.14 & 74.2 &  & \multicolumn{2}{c|}{} &  &  &  \\ \hline

\end{tabular}
}
    \caption{Classifying Modal Events: RoBERTa Results (ambiguities not shown). We unified wishes and goals into \textit{intentions} for reasons of data sparsity.}
    \label{tab:roberta}
\end{table*}

\end{document}